\documentclass[runningheads]{llncs}
\usepackage{amsmath}
\usepackage{graphicx}
\usepackage{algorithm}
\usepackage{algorithmic}
\begin{document}
\title{A Temporal Knowledge Graph Completion Method Based on Balanced Timestamp Distribution}
\titlerunning{ }

\author{Kangzheng Liu\inst{1} \and
Yuhong Zhang\inst{2}}

\authorrunning{ }

\institute{Huazhong University of Science and Technology, Wuhan, 430074, China \and School of Computer Science and Information Engineering, Hefei University of Technology, Hefei, China\\
\email{frankluis@hust.edu.cn}
\email{zhangyh@hfut.edu.cn}}

\maketitle              

\begin{abstract}
Completion through the embedding representation of the knowledge graph (KGE) has been a research hotspot in recent years. Realistic knowledge graphs are mostly related to time, while most of the existing KGE algorithms ignore the time information. A few existing methods directly or indirectly encode the time information, ignoring the balance of timestamp distribution, which greatly limits the performance of temporal knowledge graph completion (KGC). In this paper, a temporal KGC method is proposed based on the direct encoding time information framework, and a given time slice is treated as the finest granularity for balanced timestamp distribution. A large number of experiments on temporal knowledge graph datasets extracted from the real world demonstrate the effectiveness of our method.
\end{abstract}
\section{Introduction}
The knowledge graph was proposed by Google in 2012 in order to optimize the results of search engines. Many KGs including Freebase~\cite{ref_proc4}, Richpedia~\cite{ref_proc2} are created, then widely used in various applications, such as knowledge Q\&A~\cite{ref_proc5}, recommendation system~\cite{ref_proc7}, etc.

There are two kinds of KG, static KG and dynamic KG. Most dynamic KGs are related with time, which are also called temporal knowledge graph. Static KGs have a common limitation that no matter when the facts remain unchanged. Compared with static KGs (which is represented as triples $(head, relation, tail)$),  the temporal KGs are expanded into quadruples $(head, relation, tail, time)$, and the time information can effectively constrain the geometric structure of the vector space. Static KGC ignores the time information of triples. Only a small amount of existing works have made certain explorations in the field of temporal KGC. Jiang~\cite{ref_proc11} learns the time sequence between relations as a condition for the knowledge graph completion. Dasgupta~\cite{ref_proc12} directly encodes time information by projecting entities and relations onto time-specific hyperplanes. However, the distribution of timestamps in the knowledge graph is unbalanced, and each fact is projected to all the timestamps included in its time span, which results in the “overlapping part timestamps” between different facts for the same completion task. The unbalanced distribution of timestamps will aggravate this problem.

For example, given a series of temporal knowledge graph completion tasks $(Charles\_Darwin,hasWonPrize,Royal\_Medal,[1853,3000])$, $(Arthur\_Smith$-$\_Woodward,hasWonPrize,Royal\_Medal,[1917,3000])$, $(Hans\_Adolf\_Krebs,h$-$asWonPrize,Royal\_Medal,[1954,3000])$, where the end time of triples is replaced with the default maximum value due to being unknown, there will be multiple triples with the same relations and tail entities to be projected onto the same timestamp hyperplane in the time of “overlapping part”. And the training of the head entity will be indistinguishable. When calculating the score for head prediction, the parameter Dasgupta~\cite{ref_proc12} feeds is the number of the timestamp hyperplane where the start time (such as 1853, 1917, 1954) of the triples is located. But it is obvious that a triple is valid for a range of timestamps. The purpose of this is to avoid the above influence as much as possible, because the timestamp of the start time is less likely to have confused triples. The prediction of the earlier triples will be more accurate, and the prediction of the later triples will be more ambiguous, because the timestamp hyperplane, in which the later triple's start year is located, will contain more confused triples. Although Tang~\cite{ref_proc13} tries to solve this problem, it does not directly model in the embeddings to distribute the timestamps evenly. Instead, it uses an external unit (design a timespan gate in GRU~\cite{ref_proc31}) to solve this problem, which greatly increases the complexity of the model.

In order to distribute balanced timestamps in the process of modeling in the knowledge graph embedding, we proposed bt-HyTE (Hyperplane-based Temporally aware KG Embedding with Balanced timestamp distribution) and tr-HyTE (Hyperplane-based Temporally aware KG Embedding with Balanced timestamp distribution and Negative Sampling for Temporal Relations). Both use a certain time (such as year) as the finest granularity to count the number of facts in the time to ensure that the number of facts in each distributed timestamp is within the set number threshold. And then the two models project the head entities, tail entities and relations of each fact to the corresponding timestamp hyperplanes, and learn the embedding of the entities, relations, and the normal vectors of timestamp hyperplanes. Lastly, tr-HyTE adds negative sampling for temporal relations on the basis of bt-HyTE. The following is the main contributions of our paper:

\begin{itemize}
\item[$\bullet$] We propose the method bt-HyTE to generate the knowledge graph vector space. Different from the previous methods, it directly solves the problem of unbalanced timestamp distribution in the embedding of the model. Taking a certain time width as the finest granularity, bt-HyTE sequentially accumulates the number of facts in each finest granularity unit, then sets a threshold for the number of facts to have a balanced distribution for the timestamps.
\item[$\bullet$] Following bt-HyTE, we design and add negative sampling for the temporal relations to optimize the vector space of the temporal knowledge graph, and propose tr-HyTE.
\item[$\bullet$] Through a large number of experiments on real-world  datasets rich in time information, we have proved the effectiveness of bt-HyTE and tr-HyTE on temporal KGC.
\end{itemize}

The rest of this paper is organized as follows. Section 2 introduces related work, including various existing static and temporal KGC methods. The proposed method is detailed in Section 3. And the experimental details are presented in Section 4. The results are analyzed in Section 5, followed by our conclusion and future work in the final section.

\section{Related Work}
KGC is a downstream task of KGE. It completes the missing entities or relations by predicting.

The most popular methods are translation-based. The basic idea of the model TransE~\cite{ref_proc14} proposed in 2013 is that the relation is regarded as a translation from the head to the tail, which is simple and efficient. TransH~\cite{ref_proc15} is proposed to overcome the shortcomings of TransE in solving many-to-one and one-to-many relations by projecting entities on relation-specific hyperplanes. TransR~\cite{ref_proc16} is proposed based on TransE and TransH designing a relation-specific space. The translation-based models also includes TransD~\cite{ref_proc17} and TransF~\cite{ref_proc18}. In addition, there are methods based on matrix decomposition, including RESCAL~\cite{ref_proc19}, DistMult~\cite{ref_proc20}, HolE~\cite{ref_proc21}, ComplEX~\cite{ref_proc22} and SimplE~\cite{ref_proc23}. Based on the Tucker decomposition of binary tensors, a new linear model TuckER~\cite{ref_proc24} is proposed. There are some nonlinear models, such as ConvE~\cite{ref_proc25} and GNNs~\cite{ref_proc26} that have achieved high results at present, and ConvKB~\cite{ref_proc27} and CapsE~\cite{ref_proc28} have been questioned due to the work of Sun~\cite{ref_proc29}.

Although these models have good results in the experiments and datasets they show, they all ignore the time information of the knowledge graph. There are few explorations about the completion methods of temporal knowledge graphs. Jiang~\cite{ref_proc11} provides a link prediction strategy by modeling the sequence of temporal relations, and Leblay~\cite{ref_proc30} studies various methods of the interaction between time and relation through learning the embedding of them. Dasgupta~\cite{ref_proc12} proposes a model directly encoding time information in the embedding, which gives us great enlightenment. But the problem of unbalanced timestamp distribution has not been wholly solved in these embedding methods. Later, TDG2E~\cite{ref_proc13} proposed in 2020 took into account the evolution effect of time on the basis of the HyTE~\cite{ref_proc12}. Although it has considered unbalanced distribution of timestamps, the problem is solved by designing a timespan gate in GRU~\cite{ref_proc31} instead of directly embedding in the model. However, we directly and evenly distribute the timestamps in the model embedding, which not only ensures the model effectiveness but also greatly reduces the complexity of the model.

\section{Method}
In this section, we describe in detail the proposed method which consists of bt-HyTE and tr-HyTE. bt-HyTE directly handles the problem of unbalanced distribution of timestamps in model embedding. And tr-HyTE adds negative sampling for temporal relations on the basis of bt-HyTE. We first describe the overall architecture of our proposed method in Section 3.1. Next, we detail the proposed model in Section 3.2, Section 3.3, and Section 3.4.

\subsection{Overall Architecture}
Given an input knowledge graph $\mathcal{G}$ containing time information, in which the fact can be expressed in the form of a quadruple $(h,r,t,[t_s, t_e])$, we aim to complete the missing entities and relations under the condition of time. Similar to the existing model, we represent the head entity as $h$, the tail entity as $t$, and the relation as $r$, where $h,r,t \in \mathcal{R}^{d \times 1}$, $d$ is the dimension of the embedding vector.

The overall architecture of our proposed method will consist of the following three steps:
\begin{itemize}
\item[$\bullet$] Balanced distribution of the timestamps is completed by proposed balanced timestamp distribution method in the embedding of the model, which is detailed in Section 3.2.
\item[$\bullet$] Based on bt-HyTE and tr-HyTE, the entities and relations are projected onto all the time-specific hyperplanes covered by a particular triple, and then the embedding of entities and relations is learned on all the projected hyperplanes. Learning temporal knowledge graph embedding representation is described in Section 3.3.
\item[$\bullet$] To complete entity and relation, we compute the scores for each triple in the test set. And for a triple, the scores of elements in all entity set or relation set are arranged in the ascending order. And then the effectiveness of the temporal KGC is evaluated based on the evaluation metrics. See Section 3.4 for details.
\end{itemize}

\subsection{Balanced Distribution Method For Timestamps}
The quadruple $(h,r,t,[t_s, t_e])$ of temporal knowledge graph $\mathcal{G}$ indicates that the triple $(h,r,t)$ is valid in the period $[t_s,t_e]$. Assuming that the timestamps are distributed in some way, a fact is valid on the timestamp spans occupied by its time range $[t_s,t_e]$. So the rationality of the timestamp distribution will directly influence the performance of the method. For distributing $N$ timestamps, the dynamic temporal knowledge graph $\mathcal{G}$ will be partition into $N$ static knowledge graphs $\mathcal{G}_{i}$, where $i\in\{1,2,\cdots,N\}$:$\mathcal{G}=\mathcal{G}_{1} \cup \mathcal{G}_{2} \cup \cdots \mathcal{G}_{t} \cdots \cup \mathcal{G}_{N}$.

\begin{algorithm}[htb]
\caption{ Fine-grained balanced timestamp distribution.}
\label{alg:Algorithm}
\begin{algorithmic}[1] 
\REQUIRE ~~\\ 
    The time information set of all facts in the temporal knowledge graph dataset containing $N$ facts, $T=\{[t_{s_1},t_{e_1}],[t_{s_2},t_{e_2}],\cdots,[t_{s_N},t_{e_N}] \}$;\\
    The most fine-grained timeslice width, $S$;\\
    The threshold of the number of facts that distribute the timestamps, $THR$;
\ENSURE ~~\\ 
    balanced distribution set of timestamps, $T_{timestamps}$;
    \STATE Initialize a list of fact times $fact\_time$, a list of timestamps $time\_class$ and a dictionary of int $time\_freq$;
    \FOR {each $[t_{s_i},t_{e_i}]$ in T}
	 \FOR {$t=t_{s_i}$; $t \leq t_{e_i}$; $t+=S$}
	 \STATE put t into the list $fact\_time$;
	 \ENDFOR
    \ENDFOR
    \STATE $n=\lceil (max(fact\_time)-min(fact\_time))/S \rceil$;
    \STATE In temporal order, count the frequency of occurrence of each time in $fact\_time$ to the dictionary $time\_freq=\{ft_1:freq_1,ft_2:freq_2,\cdots,ft_n:freq_n\}$;
    \FOR {each $\{ft_i:freq_i\}$ in $time\_freq$}
    \STATE count += $freq_i$;
	 \IF {count > $THR$}
	 \STATE put $ft_i$ into the list $time\_class$;
	 \ENDIF
	 \ENDFOR
    \STATE Complete the generation of the timestamp set $T_{timestamps}$ through $time\_class$;
    \STATE return $T_{timestamps}$; 
\end{algorithmic}
\end{algorithm}

We consider dealing with timestamp distributing directly in the embedding of the model to mitigate the effects of the problems described above. We take a period of time as the the finest granularity, and collect the most fine-grained time period for each fact, in which the fact is valid. If a year is used as the most fine-grained time period, the triple $(h,r,t)$ is valid in every year within $[t_s, t_e]$, where $\{t|t_s\leq t \leq t_e,t \  is\ the\ most\ fine\_grained\ time\ slice\}$ , which is included in the set of effective time spans. Then, the number of facts contained in each of the most fine-grained times is counted in the embedding, and the threshold for distributing the timestamps is set, thereby the balanced distribution of the timestamps is done. As shown in Alg. 1, we consider not only the start time and the end time of a fact, but also the process time between them, thus taking into account all the times when the fact is valid. For the $(-,hasWonPrize,Royal\_Medal)$ series triple illustrated above, according to the number of facts of the most fine-grained time slice, the later time will accumulate more fact numbers as time advances, and then when the threshold is reached, it is able to be distributed into $time\_class$, by which we distribute the timestamps. With this newly formed timestamp, the prediction of missing entities or relations prior to the year will be conditionally completed.

\begin{figure}
\begin{center}
\includegraphics[width=0.50\textwidth]{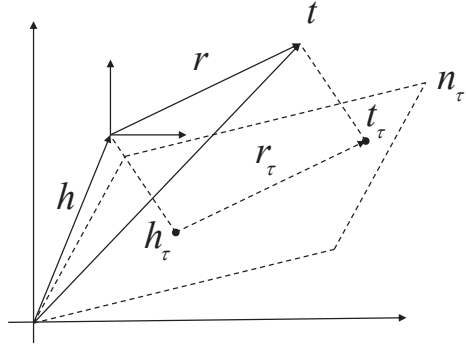}
\caption{In the figure, vectors of a triple $h$, $r$, $t$ are projected onto the time-specific hyperplane $n_{\tau}$, where the triple $(h,r,t)$ is valid.} \label{fig1}
\end{center}
\end{figure}
\subsection{Temporal Knowledge Graph Embedding Representation}
In the above temporal knowledge graph $\mathcal{G}$, $\mathcal{T}$ is the set of timestamps distributed in the embedding of the model, then the set of timestamps covered by the time period $[t_s,t_e]$ is expressed as $\{\tau|\tau\in\mathcal{T}\}$. Project entity and relation vectors onto the hyperplane of a particular timestamp $n_{\tau}$ under the balanced distribution of timestamps as shown in Fig.~\ref{fig1},
\begin{equation}
\begin{array}{l}
P\left(h,\tau\right)=h-\left(n_{\tau}^{\top} h\right) n_{\tau} \\
P\left(t,\tau\right)=t-\left(n_{\tau}^{\top} t\right) n_{\tau} \\
P\left(r,\tau\right)=r-\left(n_{\tau}^{\top} r\right) n_{\tau}
\end{array}
\end{equation}

\noindent The score function is expressed as:
\begin{equation}
f(h, r, t,\tau)=\left\|P\left(h,\tau\right)+P\left(r,\tau\right)-P\left(t,\tau\right)\right\|_{l_{1} / l_{2}}
\end{equation}

For a valid fact, there should be $P(h,\tau)+P(r,\tau)\approx P(t,\tau)$, where $\tau$ is a timestamp evenly distributed in the embedding of the model using Alg. 1. For a fact triple in a negative sample, $|P\left(h,\tau\right)+P\left(r,\tau\right)-P\left(t,\tau\right)|$ should be a number much greater than 0. Here, we learn the normal vector of each timestamp hyperplane, and the embedding of projected entities and relations on each timestamp hyperplane.

\noindent We learn the model by minimizing the loss function, which is expressed as:
\begin{equation}
Loss=\sum_{\tau=1}^{N} \sum_{x \in \mathcal{D}^{+}} \sum_{y \in \mathcal{D}^{-}} \max \left(0, f(x,\tau)-f(y,\tau)+\gamma\right)
\end{equation}
As mentioned, for a corrupted triple, the distance should be a very large positive number, which will result in $Loss$ is less than 0. So for a triple, when its $Loss$ is greater than 0, the model need to continue to learn. $\mathcal{D}^{+}$ is the set of valid triples. $\mathcal{D}^{-}$ is the set of negative samples, and we generate negative samples in following two ways:
\begin{itemize}
\item[$\bullet$] {\bfseries Without negative sampling for the temporal relations}.
A certain number of negative sampling is carried out for the head and tail entities of valid triples without considering the influence of the timestamps, and the negative sampling strategy is applied to bt-HyTE expressed by the following formula,

\begin{small}$\begin{array}{r}
\mathcal{D}^{-}=\{(h^{\prime}, r, t, \tau), (h, r, t^{\prime}, \tau) \mid h^{\prime} \in \mathcal{E}, t^{\prime} \in \mathcal{E}, (h,r,t) \in \mathcal{D}^{+}, (h^{\prime}, r, t) \notin \mathcal{D}^{+},\\
(h, r, t^{\prime}) \notin \mathcal{D}^{+}\}
\end{array}$
\end{small}
\item[$\bullet$] {\bfseries With negative sampling for the temporal relations}.
A certain number of negative samples are taken for the head entities, tail entities and temporal relations of the valid triples.This negative sampling strategy is applied to tr-HyTE, and it is expressed by the following formula,

\begin{small}$\begin{array}{r}
\mathcal{D}^{-}=\{(h^{\prime}, r, t, \tau), (h, r^{\prime}, t, \tau), (h, r, t^{\prime}, \tau) \mid h^{\prime} \in \mathcal{E}, r^{\prime} \in \mathcal{R}, t^{\prime} \in \mathcal{E}, (h,r,t) \in \mathcal{D}^{+}, \\ (h^{\prime}, r, t) \notin \mathcal{D}^{+}, (h,r^{\prime},t) \notin \mathcal{D}^{+}, (h, r, t^{\prime}) \notin \mathcal{D}^{+}\}
\end{array}$
\end{small}
\end{itemize}

We performed link prediction, relation prediction tasks, and some fine-grained investigation to prove the effectiveness of proposed bt-HyTE and tr-HyTE. The addition of negative sampling strategy for temporal relations can greatly improve the performance of relation prediction, but it can also inhibit the performance of head and tail entity prediction to a certain extent.

\subsection{Temporal Knowledge Graph Completion}
The temporal KGC is entity prediction and relation prediction tasks based on the embedding of entities, relations, and times. At present, the widely accepted task of temporal KGC takes time as the condition of the traditional completion task. We also use time as the condition of entity prediction and relation prediction to further improve the performance of KGC. In this paper, the temporal KGC task is divided into the following subtasks:

\begin{itemize}
\item[$\bullet$] {\bfseries Head prediction}: for an incomplete fact $(?,r,t,\tau)$, predict the head entity $h$ from the entity set at period $\tau$.
\item[$\bullet$] {\bfseries Relation prediction}: for an incomplete fact $(h,?,t,\tau)$, predict the relation $r$ from the relation set at period $\tau$.
\item[$\bullet$] {\bfseries Tail prediction}: for an incomplete fact $(h,r,?,\tau)$, predict the tail entity $t$ from the entity set at period $\tau$. 
\end{itemize}

\section{Experiments}
In this section, we experiment bt-HyTE and tr-HyTE based on popular datasets and compare their performance with the state-of-the-art KGC methods. Specifically, we first introduce the datasets used, the methods involved in the comparison, and the evaluation schema, and then introduce the details of the experiment.
\subsection{Datasets}
We use the subgraphs containing rich time information of Wikidata and YAGO knowledge graph datasets to carry out the experiment. The subjects of the experiment included bt-HyTE, tr-HyTE and some baselines.
\begin{itemize}
\item[$\bullet$] {\bfseries YAGO11k}. This is a temporal knowledge graph from YAGO3~\cite{ref_proc32}. The extracting method was proposed by HyTE~\cite{ref_proc12}. It consists of 20.5k triples, 10,623 entities, and 10 relations.
\item[$\bullet$] {\bfseries Wikidata11k}. This is a temporal knowledge graph extracted from Wikidata by the method proposed in HyTE~\cite{ref_proc12}. It contains 27.5k triples, 10623 entities and 24 relations.
\end{itemize}
\begin{table}
\begin{center}
\caption{Details of the YAGO11k dataset and Wikidata11k dataset.}\label{tab1}
\begin{tabular}{|c|c|c|c|c|c|}
\hline
Datasets & \#Entity & \#Relation & \#Train & \#Valid & \#Test\\
\hline
YAGO11k & 10623 & 10 & 16.4k & 2k & 2k\\
\hline
Wikidata11k & 10623 & 24 & 23.4k & 2k & 2k\\
\hline
\end{tabular}
\end{center}
\begin{center}
\caption{Optimal parameter configuration of bt-HyTE and tr-HyTE. lr denotes learning rate, testfreq denotes the frequency of calculating the score of valid set in each epoch, inpdim denotes dimension of embedded vector, margin denotes the distance between positive and negative samples, l2 indicates whether l2 distance is used, and we use 0, so that l1 distance is applied to calculate the score, neg\_sample denotes the number of negative samples for each positive sample.}\label{tab2}
\begin{tabular}{|c|c|c|c|c|c|c|}
\hline
Datasets & lr & testfreq & inpdim & margin & l2 & neg\_sample\\
\hline
YAGO11k & 0.0001 & 5 & 128 & 10 & 0 & 5\\
\hline
Wikidata11k & 0.0001 & 5 & 128 & 10 & 0 & 5\\
\hline
\end{tabular}
\end{center}
\end{table}

\subsection{Compared Methods}
As following, we compare these five methods to evaluate the performance of our method:
\begin{itemize}
\item[$\bullet$] {\bfseries TransE}~\cite{ref_proc14}: TransE is the pioneering work of translation-based models, which embeds entities and relations in the same vector space. It is simple but efficient.
\item[$\bullet$] {\bfseries TransH}~\cite{ref_proc15}: TransH put forward on the basis of TransE. It projects entities onto the relation-specific hyperplane, and regards the relation on the hyperplane as the translation from the head to the tail.
\item[$\bullet$] {\bfseries HyTE}~\cite{ref_proc12}: HyTE is a dynamic KGE method. Inspired by TransH's modeling, it projects entities and relations onto the time-specific hyperplane, and regards the projected relation as the translation of the projected head to the projected tail. It has achieved remarkable results in dealing with the completion of temporal KGs.
\item[$\bullet$] {\bfseries bt-HyTE}: On the basis of HyTE, we consider directly dealing with unbalanced timestamp distribution in the model embedding, and propose bt-HyTE.
\item[$\bullet$] {\bfseries tr-HyTE}: On the basis of bt-HyTE, we add negative sampling for the temporal relations, and the model is named tr-HyTE, which effectively improves the performance of relation prediction.
\end{itemize}

\subsection{Evaluation Schema}
We use the traditional evaluation metrics Mean Rank and HIT@k in the field of KGC. The vectors embedded in the low-dimensional space are first projected onto the time-specific hyperplanes by formula (1), and then the scores are calculated and ordered by formula (2). The rankings of the missing entities are recorded, and the following metrics are calculated:
\begin{itemize}
\item[$\bullet$] {\bfseries Mean Rank}: For each triple in the test set, scores calculated are sorted from small to large, then the missing entity or relation ranking of each triple is counted. The average value of rankings of all the triples in test set is called Mean Rank. For Mean Rank, a smaller value is better.
\item[$\bullet$] {\bfseries HIT@k}: For each triple in the test set, scores calculated are also sorted from small to large, and the proportion of triples whose missing entity or relation is ranking in front of k to the total number of the test sets is counted. For HIT@k, a larger value is better.
\end{itemize}

\subsection{Experiment Details}
\subsubsection{Experiment Setup} For all the methods, the max epoch during training is 4000, and we adopt the optimal parameter configuration given in the corresponding method's paper. The dimension of the embedding vector is selected from $\{100,128,256\}$ and takes 128; Margin is selected from $\{1,5,10\}$ using 10; Optimizer takes SGD; Learning rates selected from $\{0.05,0.001,0.0001\}$ are all 0.0001; And batch size will be selected from $\{2000,5000,50000\}$; Neg sample is selected from $\{1,2,5\}$. The optimal parameter configuration of the proposed method is shown in Table~\ref{tab2}.
\subsubsection{Entity Prediction} Entity prediction includes two subtasks: head prediction and tail prediction. Like previous models, the selection of hyperparameter epoch is also dependent on the mean of Mean Rank metric of head prediction and tail prediction. For the fact $(h,r,t,\tau)$, the head entity prediction is to process the case of $ (?,r,t,\tau)$, and the tail entity prediction is to process the case of $(h,r,?,\tau)$, both of which regard time $\tau$ as a known condition of completion. We have experimented with baselines, bt-HyTE and tr-HyTE respectively.
\subsubsection{Relation Prediction} The selection of the hyperparameter epoch of the relation prediction directly depends on the Mean Rank metric of the relation prediction. Different from the static KGE methods, the relation prediction in the temporal KGs is to process the case of $(h,?,t,\tau)$. Knowing the head $h$ and tail $t$, the relation prediction for the triple is completed under the condition of time $\tau$. We also experiment on the relation prediction of bt-HyTE, tr-HyTE and some baselines.
\subsubsection{Fine-grained Investigation Based On YAGO11k Dataset} Based on the YAGO11k test set with realistic representational significance and HIT@k metric, we fine-grain comparisons of completion performance between different methods. The completion strategy is as follows:
\begin{itemize}
\item[$\bullet$] {\bfseries Relation completion}: The metric referenced is HIT@1, which directly completes the number one relation and forms the valid facts. 
\item[$\bullet$] {\bfseries Head \& tail completion}: The metric referenced is HIT@10. If the score of the missing entity is in the top 10, then it is defined that it can complete the missing entity and form the valid fact.
\end{itemize}

\section{Results And Analysis}
To verify the strengths of our method, we not only make statistics and analysis of the results based on Mean Rank and HIT@k metrics, but also have a fine-grained investigation on bt-HyTE, tr-HyTE and compared models.

\subsection{Performance Analysis And Comparison}
\begin{table}
\begin{center}
\caption{Comparison between different methods based on Wikidata11k dataset.}\label{tab3}
\begin{tabular}{|c|c|c|c|c|c|c|}
\hline
Datasets & \multicolumn{6}{|c|}{Wikidata11k}\\
\hline
Metric & \multicolumn{3}{|c|}{MR} & \multicolumn{2}{|c|}{HIT@10(\%)} & HIT@1(\%)\\
\hline
Tasks & head & tail & relation & head & tail & relation\\
\hline
TransE & 900.72 & 694.49 & 2.81 & 8.31 & 23.02 & 75.81\\
TransH & 867.87 & 653.63 & 2.71 & 9.19 & 26.74 & 72.73\\
HyTE & {\bfseries 448.28} & 308.91 & 1.55 & 28.40 & 48.58 & 81.28\\
\hline
bt-HyTE & 448.71 & {\bfseries 282.90} & 1.51 & 28.79 & 50.34 & 81.52\\
tr-HyTE & 450.21 & 318.62 & {\bfseries 1.30} & {\bfseries 30.21} & {\bfseries 50.49} & {\bfseries 86.46}\\
\hline
\end{tabular}
\end{center}
\begin{center}
\caption{Comparison between different methods based YAGO11k dataset.}\label{tab4}
\begin{tabular}{|c|c|c|c|c|c|c|}
\hline
Datasets & \multicolumn{6}{|c|}{YAGO11k}\\
\hline
Metric & \multicolumn{3}{|c|}{MR} & \multicolumn{2}{|c|}{HIT@10(\%)} & HIT@1(\%)\\
\hline
Tasks & head & tail & relation & head & tail & relation\\
\hline
TransE & 2042.04 & 447.51 & 1.47 & 1.27 & 7.95 & 78.11\\
TransH & 1891.57 & 351.83 & 1.48 & 1.32 & 11.70 & 79.52\\
HyTE & 1059.62 & {\bfseries 110.53} & 1.28 & 13.31 & 35.01 & 82.64\\
\hline
bt-HyTE & 1040.78 & 114.07 & 1.27 & 13.46 & {\bfseries 37.06} & 82.45\\
tr-HyTE & {\bfseries 1036.90} & 118.99 & {\bfseries 1.22} & {\bfseries 14.92} & 34.67 & {\bfseries 85.18}\\
\hline
\end{tabular}
\end{center}
\end{table}
Our proposed models bt-HyTE and tr-HyTE are better than other compared models in performance, which may be due to the effectiveness of our contributions.

As shown in Table~\ref{tab3} and Table~\ref{tab4}, the best results are bolded. bt-HyTE are qualitatively improved compared to not only the static KGC models TransE and TransH, but also the current state-of-the-art temporal knowledge graph embedding model HyTE. Based on Wikidata11k dataset, bt-HyTE is better than HyTE on MR of tail prediction (an increase of 26.01) and relation prediction (an increase of 0.04). As for HIT@k metric, bt-HyTE has an increase of 0.39\% on head prediction, 1.76\% on tail prediction and 0.24\% on relation prediction compared with HyTE. Based on YAGO11k dataset, bt-HyTE is better than HyTE on MR of head prediction, relation prediction and HIT@k of head prediction, tail prediction. There is a slight fluctuation in other metrics, which may be due to the experimental error. It illustrates the effectiveness of balanced timestamps.

Compared with bt-HyTE, tr-HyTE has further improved performance. Based on YAGO11k dataset, tr-HyTE is better than bt-HyTE on MR of head prediction (an increase of 3.88) and relation prediction (an increase of 0.05). In HIT@k metrics, tr-HyTE has an increase of 1.46\% on head prediction and 2.73\% on relation prediction. Based on Wikidata11k dataset, tr-HyTE is also better than bt-HyTE on MR of relation prediction and HIT@k of head prediction, tail prediction, relation prediction. And tr-HyTE is obviously better than HyTE. It demonstrates the effectiveness of balanced timestamps again and illustrates the effectiveness of negative sampling for temporal relations.

However, we also notice that MR-based head entity prediction on the Wikidata11k dataset and MR-based tail entity prediction on the YAGO11k dataset are not very well. We attribute this to the problem that a certain year contains a very large number of facts. Although our timestamp fact number threshold has been set, the number of facts included in some of the finest-grained time slices is much more than the threshold, which obviously weakens the balance of timestamp distribution.

\subsection{Fine-grained Investigation Results}

\begin{table}
\begin{center}
\caption{Fine-grained investigation for head prediction based on YAGO11k dataset. The missing entity that hits the top ten is bolded, and the missing entity that does not hit the top ten is displayed as a horizontal bar.}\label{tab5}
\resizebox{\textwidth}{!}{
\begin{tabular}{|c|c|c|c|c|}
\hline
Test Quadruples & TransE & HyTE & bt-HyTE & tr-HyTE\\
\hline
?, playsFor, France\_national\_football\_team, [2002, 2011] & - & - & {\bfseries Djibril\_Cissé} & {\bfseries Djibril\_Cissé}\\
\hline
?, playsFor, Bradford\_City\_A.F.C., [2011, 2014] & - & - & {\bfseries Kyel\_Reid} & -\\
\hline
?, playsFor, Woking\_F.C., [2003, 2004] & - & - & - & {\bfseries Ashley\_Bayes}\\
\hline
?, hasWonPrize, Eisner\_Award, [2001, 2010] & - & - & - & {\bfseries J.\_H.\_Williams\_III}\\
\hline
\end{tabular}}
\end{center}
\begin{center}
\caption{Fine-grained investigation for tail prediction based on YAGO11k dataset. The missing entity that hits the top ten is bolded, and the missing entity that does not hit the top ten is displayed as a horizontal bar.}\label{tab6}
\resizebox{\textwidth}{!}{
\begin{tabular}{|c|c|c|c|c|}
\hline
Test Quadruples & TransE & HyTE & bt-HyTE & tr-HyTE\\
\hline
Alex\_Campbell\_(golfer), wasBornIn, ?, [1876, 1876] & - & - & - & {\bfseries Scotland}\\
\hline
Sarney\_Filho, wasBornIn, ?, [1957, 1957] & - & - & {\bfseries São\_Luís,\_Maranhão} & {\bfseries São\_Luís,\_Maranhão}\\
\hline
Marion\_Burns, wasBornIn, ?, [1907, 1907] & - & - & {\bfseries Los\_Angeles} & -\\
\hline
Brad\_Mays, wasBornIn, ?, [1955, 1955] & - & - & - & {\bfseries St.\_Louis}\\
\hline
\end{tabular}}
\end{center}
\begin{center}
\caption{Fine-grained investigation for relation prediction based on YAGO11k dataset. Relations are displayed in the order of completion. The correct prediction is in bold.}\label{tab7}
\resizebox{\textwidth}{!}{
\begin{tabular}{|c|c|c|c|c|}
\hline
Test Quadruples & TransE & HyTE & bt-HyTE & tr-HyTE\\
\hline
Tomoko\_Kawase, ?, Kyoto, [1975, 1975] & diedIn, isMarriedTo & diedIn, isMarriedTo & diedIn, isMarriedTo & {\bfseries wasBornIn}, diedIn\\
\hline
Sacha\_Guitry, ?, Saint\_Petersburg, [1885, 1885] & diedIn, {\bfseries wasBornIn} & diedIn, created & diedIn, created & {\bfseries wasBornIn}, diedIn\\
\hline
Eleanor\_Summerfield, ?, London, [1921, 1921] & diedIn, {\bfseries wasBornIn} & diedIn, {\bfseries wasBornIn} & {\bfseries wasBornIn}, diedIn & diedIn, {\bfseries wasBornIn}\\
\hline
Olivia\_Newton-John, ?, Cambridge, [1948, 1948] & diedIn, isMarriedTo & diedIn, {\bfseries wasBornIn} & diedIn, {\bfseries wasBornIn} & {\bfseries wasBornIn}, diedIn\\
\hline
\end{tabular}}
\end{center}
\end{table}

Through the fine-grained investigation method described in Section 4.4, we select a certain number of specific quadruples based on the test set of the YAGO11k dataset, and prove the effectiveness of our proposed model from the fine-grained perspective.

As shown in Table~\ref{tab5} and Table~\ref{tab6}, on the entity prediction task, our proposed bt-HyTE and tr-HyTE can complete the missing entities that rank top 10 ahead of the static KGC method TransE and the temporal model HyTE that ignores the balanced distribution of timestamps. As shown in Table~\ref{tab7}, in the relation prediction task, for some specific quadruples, HyTE confuses the similar relations between “dieIn” and “wasborn” like TransE, but our proposed bt-HyTE and tr-HyTE can distinguish these relations. The performance of tr-HyTE is better than that of bt-HyTE for relation prediction, which is because of the negative sampling for temporal relations. But in our method, the number of negative samples for temporal relations should be strictly controlled to avoid affecting the entity prediction.

\section{Conclusion}
In this paper, we proposed a temporal KGC method bt-HyTE, and on this basis, a method of adding negative sampling for temporal relations was proposed named tr-HyTE. Based on the original method of directly encoding time information HyTE, we consider directly solving the problem of unbalanced timestamp distribution in the model embedding. The experimental results based on temporal knowledge graph datasets indicate that our proposed model is better than the state-of-the-art algorithms. In the future, we plan to further refine the time slice size used for balanced timestamp distribution, and further study the influence of this time slice size on the effectiveness of the temporal KGC methods.

%
%
%
%

\end{document}